\documentclass{article}
\usepackage{ijcai24}

\usepackage{times}
\usepackage{soul}
\usepackage{url}
\usepackage[hidelinks]{hyperref}
\usepackage[utf8]{inputenc}
\usepackage[small]{caption}
\usepackage{graphicx}
\usepackage{amsmath}
\usepackage{amsthm}
\usepackage{booktabs}
\usepackage{algorithm}
\usepackage{algorithmic}
\usepackage[switch]{lineno}

\usepackage{booktabs}
\usepackage{multirow}
\usepackage{adjustbox}
\usepackage{xcolor}

\linenumbers

\title{Where Would I Go Next? Large Language Models as Human Mobility Predictors}

\author{
Xinglei Wang$^1$ $^*$ \and
Meng Fang$^2$ $^*$ \and
Zichao Zeng$^{1,3}$\And
Tao Cheng$^1$
\affiliations
$^1$SpaceTimeLab for Big Data Analytics, University College London\\
$^2$University of Liverpool\\
$^3$3DIMPact, University College London\\
\emails
\{xinglei.wang.21, zichao.zeng.21, tao.cheng\}@ucl.ac.uk,
mfang@liverpool.ac.uk
}

\begin{document}
\nolinenumbers

\maketitle

\def\thefootnote{*}\footnotetext{These authors contributed equally to this work}\def\thefootnote{\arabic{footnote}}

\begin{abstract}
    Accurate human mobility prediction underpins many important applications across a variety of domains, including epidemic modelling, transport planning, and emergency responses. Due to the sparsity of mobility data and the stochastic nature of people's daily activities, achieving precise predictions of people's locations remains a challenge. While recently developed large language models (LLMs) have demonstrated superior performance across numerous language-related tasks, their applicability to human mobility studies remains unexplored. Addressing this gap, this article delves into the potential of LLMs for human mobility prediction tasks. We introduce a novel method, \textit{LLM-Mob}, which leverages the language understanding and reasoning capabilities of LLMs for analysing human mobility data. We present concepts of \textit{historical stays} and \textit{context stays} to capture both long-term and short-term dependencies in human movement and enable time-aware prediction by using time information of the prediction target. Additionally, we design context-inclusive prompts that enable LLMs to generate more accurate predictions. Comprehensive evaluations of our method reveal that \textit{LLM-Mob} excels in providing accurate and interpretable predictions, highlighting the untapped potential of LLMs in advancing human mobility prediction techniques. The code is available at \url{https://github.com/xlwang233/LLM-Mob}
\end{abstract}

\section{Introduction}

\begin{figure}[t]
    \centering
    \includegraphics[width=0.9\columnwidth]{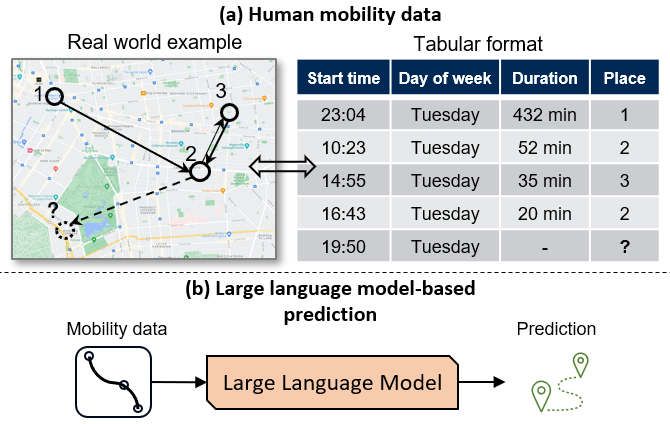}
    \caption{LLM-based human mobility prediction.}
    \label{fig:intro}
\end{figure}

Human mobility refers to the movement of people from one place to another, typically within a geographic area such as a city, region, or country. 

The prediction of human mobility has seen immediate benefits in many downstream tasks such as points of interest recommendation \cite{sanchez2022point}, communication networks traffic prediction \cite{jiang2022cellular}, road traffic optimisation 
\cite{rossi2019modelling} and many more. Moreover, studying the mobility of people is crucial for tackling numerous prominent societal challenges, including urbanisation \cite{barthelemy2019statistical}, segregation \cite{li2022aggravated}, and the spread of epidemics \cite{kraemer2020effect}, to name a few.

The unique characteristics of human mobility manifest in its inherent regularity, stochasticity \cite{song2010modelling} and complex spatiotemporal dependencies, making it hard to accurately predict people's whereabouts. Recent studies exploit the spatiotemporal modelling capabilities of deep learning models to achieve better predictive performance \cite{feng2018deepmove,xue2021mobtcast,hong2023context}, but the accuracy is still not sufficient and the produced results cannot be directly and fully explained, which hinders the interpretability and applicability of such models in real world applications.

The 
recently launched large language model (LLM) - ChatGPT \cite{openai2022chatgpt} has shown impressive performance, outperforming many models in a range of NLP tasks even in zero-shot settings \cite{qin2023chatgpt}. It has not only spawned many interesting researches in NLP field and beyond, but also profoundly changed people’s daily lives. However, whether LLMs are able to model human mobility data remains unknown. 

To answer this question, this work explores the potential of LLMs for modelling human mobility data, as shown in Figure \ref{fig:intro}. Utilising LLMs to model human mobility is not a trivial task because LLMs are trained and optimised for language processing and understanding, which means they cannot be directly used for location prediction. To address this issue, we propose a framework called \textit{LLM-Mob} that seamlessly combines human mobility prediction with language modelling. Specifically, we organise the mobility data into \textit{historical stays} and \textit{context stays} to account for long-term and short-term  dependencies in people's movements and utilise the time information from the \textit{target stay} to make time-aware prediction. Moreover, we design effective prompting strategies to help the LLMs understand the mobility data, maximise their reasoning abilities, and enable the interpretation of the prediction results. 

To validate the proposed \textit{LLM-Mob}, we conduct extensive experiments on two public human mobility datasets, in which we compare our method with the state-of-the-art models and analyse the prediction results. Experimental results show that \textit{LLM-Mob} achieves superior predictive performance and interpretability, demonstrating the promising utility of LLMs in mobility prediction. Furthermore, we probe into the capabilities of LLMs in predicting human mobility and point out the limitations of our method. 

We argue that our work constitutes a notable contribution to both human mobility and NLP research community. On one hand, we have introduced and taken the initial step towards a paradigm shift to modelling human mobility - moving from domain-specific deep neural networks to a general purpose LLM-based approach. On the other hand, we innovatively repurpose LLMs for tasks they were not originally designed for, enabling LLMs to extend beyond their traditional use cases through dedicated data organisation and prompting strategies. We envision that our work could inspire a line of research utilising LLMs in the future.


\section{Related work}

\subsection{Human Mobility Prediction}
Human mobility prediction is the study of forecasting human movements either at individual or collective levels, using various data sources and predictive modelling techniques. Several pioneering studies focused on uncovering spatiotemporal regularity and statistical properties of individuals’ movements ~\cite{gonzalez2008understanding,song2010limits}. They revealed the predictability of human mobility and laid the theoretical foundation of this research field. In recent years, with the availability of large-scale mobility data and advances in machine learning, data-driven approaches have gained popularity in human mobility prediction. Early studies assumed the Markovian property for individuals’ location visits ~\cite{ashbrook2002learning}, based on which, Mobility Markov Chain was first proposed by \cite{gambs2010show} and was later used in their work on the next place prediction \cite{gambs2012next}. Although more recent studies \cite{huang2017mining,wang2020predictability} further enhanced the predictive performance of this type of models, they failed to consider the long-range spatiotemporal dependencies. Therefore, deep learning models capable of learning complex sequential dependencies from vast amount of data were introduced to tackle the problem. For example, recurrent neural network (RNN) and its variants (e.g., LSTM \cite{hochreiter1997long} and GRU \cite{cho2014learning}) have been proven to outperform Markov models in predicting the next location \cite{feng2018deepmove,krishna2018lstm}. Moreover, most recent works \cite{xue2021mobtcast,hong2023context} have utilised Transformer architecture \cite{vaswani2017attention} to mitigate the limited ability of RNNs to model long-term dependencies and have incorporated various contextual information such as semantic, social, and geographical contexts to further improve the performance. Despite the increasing complexity of these deep neural architectures, the performance enhancements have been somewhat incremental and the prediction results cannot be fully explained. 

\subsection{Large Language Models}
Large language models (LLMs) refer to Transformer-based pre-trained language models that contain tens or even hundreds of billions of parameters and are trained on massive text data \cite{zhao2023survey}. Existing LLMs include GPT-3 \cite{brown2020language}, GPT-4 \cite{openai2023gpt4}, PaLM \cite{chowdhery2022palm}, and Llama \cite{touvron2023llama}, etc. They have not only achieved excellent performance on various language-related tasks but have also found applications outside the NLP fields such as biology, social sciences, geographic information science \cite{mai2023opportunities} and so on. The success of LLMs is closely related to their emergent abilities \cite{wei2022emergent} that are not found in small models, such as in-context learning which is briefly described below. 

\subsubsection{In-context learning.} 
The in-context learning (ICL) ability is formally introduced by GPT-3 \cite{brown2020language}: assuming that the language model has been provided with a natural language instruction and/or several task demonstrations, it can generate the expected output without requiring additional training or gradient update. Constructing effective prompts is essential to ICL. \cite{wei2022chain} proposed a prompting strategy called chain-of-thought (CoT) that can bring out the strong reasoning ability of LLMs by utilising the intermediate reasoning steps for deriving the final answer. Extensions of CoT prompting include zero-shot variants \cite{kojima2022large}, automatically generated series of reasoning steps \cite{zhang2022automatic,wang2023plan}, sampling and selecting the most consistent answer via a majority vote \cite{wang2022self}, and a tree of thoughts that enables the LLMs to self-evaluate the intermediate thoughts and incorporate search algorithms \cite{yao2023tree}. Compared to supervised learning, ICL does not require any training process, greatly reducing the computational costs for adapting the model to new tasks.

Despite the successful application in various areas, no research has touched on the possibility of using LLMs in human mobility studies. To fill the gap, we make the first attempt to leverage the cutting-edge LLMs for modelling and predicting human mobility by establishing an in-context learning process detailed in the following sections.

\section{Preliminaries}

We introduce the notations used in this article and formulate the next location prediction problem.
\subsubsection{Terms and Notations.}
\label{sec:notation} 
Mobility data are typically collected through electronic devices and stored as spatiotemporal trajectories. Each track point in a user’s trajectory comprises a pair of spatial coordinates and a timestamp. After preprocessing, a user’s trajectory is represented as a sequence of \textit{stays} where people remain stationary for a certain amount of time. More specifically, a stay is denoted as (\textit{st}, \textit{dow}, \textit{dur}, \textit{pid}), where \textit{st} indicates the time when the stay starts, \textit{dow} denotes the day of the week, \textit{dur} denotes the duration of the stay, and \textit{pid} denotes the unique identifier of the place where the stay occurred. An example of a stay could be (\textit{17:30}, \textit{Tuesday}, \textit{35 minutes}, \textit{place$\_$1}), which basically means that the user stayed at \textit{place$\_$1} from 17:30 till 18:05 on Tuesday.

\subsubsection{Next Location Prediction.} Given a user’s sequence of stays up until time $n$: ${\textbf {\textit S}}=(S_{n-Q+1},…,S_n)$, the goal is to predict the next location/place (i.e., $pid_{n+1}$) that the user will visit in the next time step. The length of the sequence \textit{Q} determines how much past information is considered in the predictive model.

\section{Methodology}

\begin{figure}[t]
    \centering
    \includegraphics[width=0.5\columnwidth]{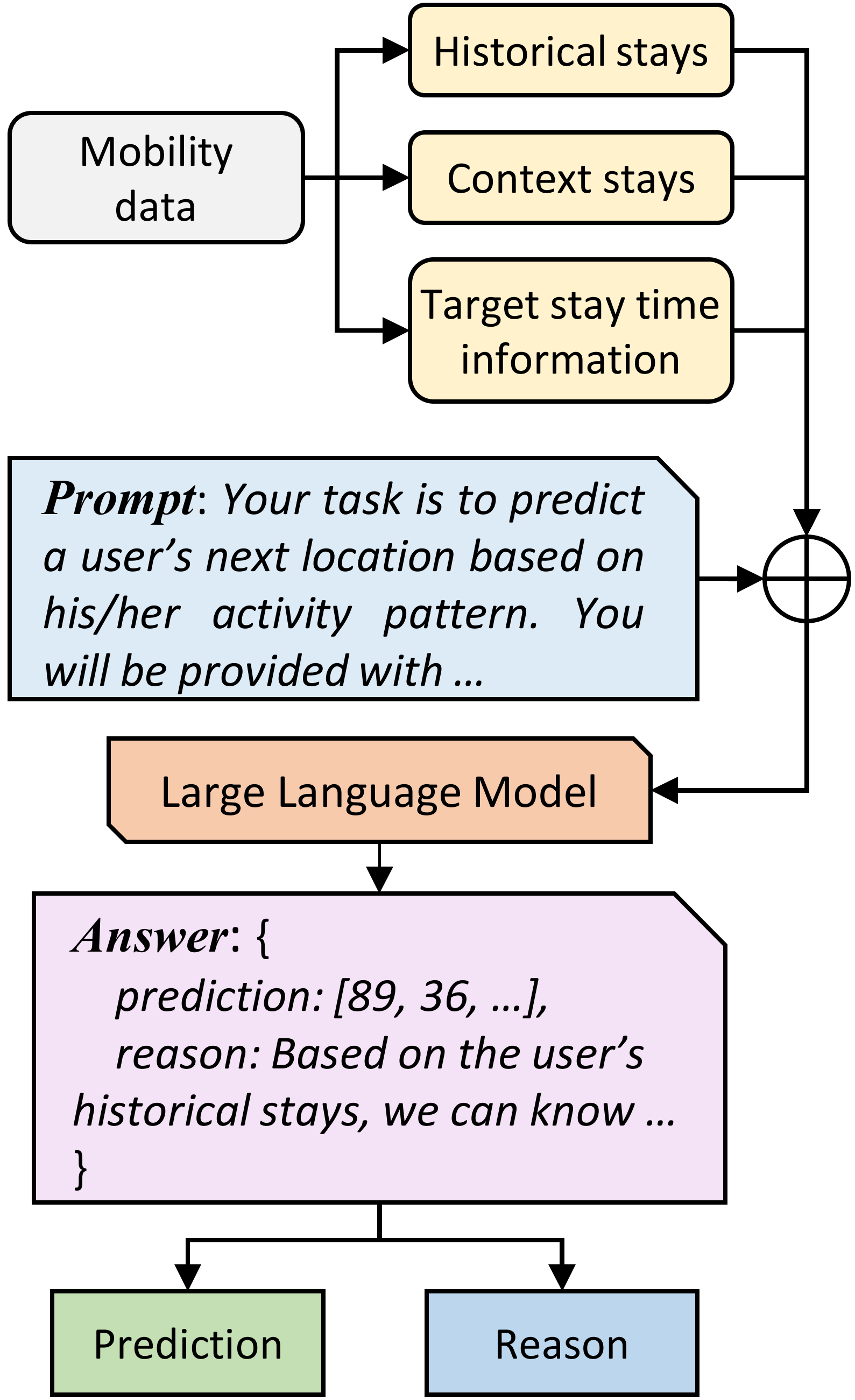}
    \caption{The workflow of \textit{LLM-Mob}.}
    \label{fig:llm-mob}
\end{figure}

In this section, we elaborate on the framework of our proposed \textit{LLM-Mob}. The whole workflow is shown in Figure \ref{fig:llm-mob}. There are two main steps: data formatting and prompt designing. In constructing the data, the pre-processed mobility data will be formatted into \textit{historical stays} and \textit{context stays} to accommodate both long-term and short-term dependencies and the time information contained in target stay, i.e., $(st_{n+1},dow_{n+1})$, is utilised to facilitate time-aware prediction. Then we use the formatted data to form \textbf{\textit{Prompt}}, which is fed into LLMs to retrieve the \textbf{\textit{Answer}} that contains both the prediction results and the corresponding reasons. The details are explained below.

\begin{figure}[t]
    \centering
    \includegraphics[width=1.0\columnwidth]{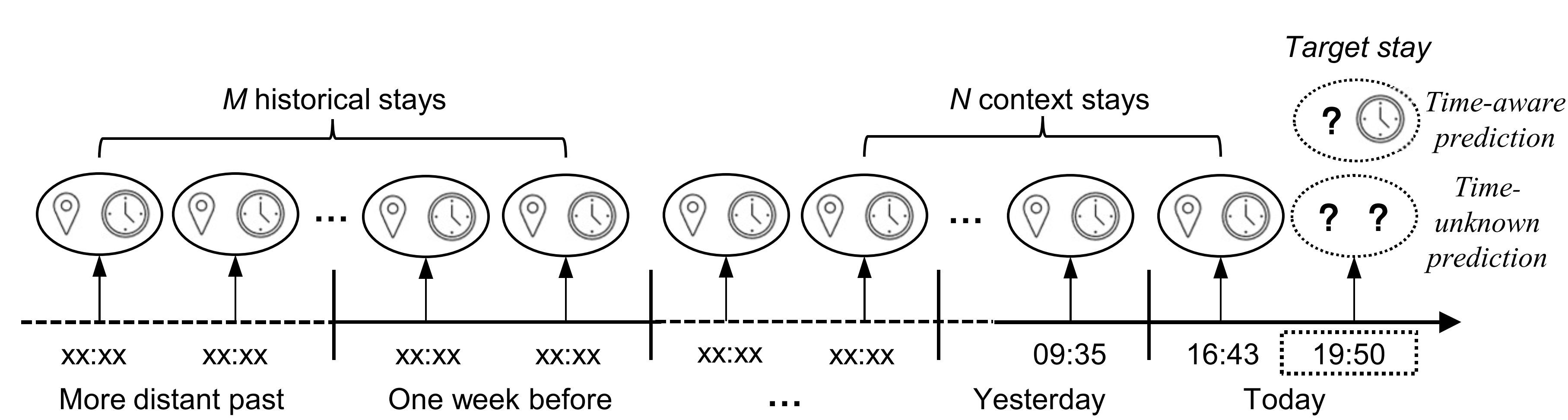}
    \caption{Illustration of historical, context and target stays.}
    \label{fig:dataorg}
\end{figure}

\subsection{Data Formatting}

It is challenging for LLMs to extract useful information from raw sequential stays, and hence we propose to format the data as specific kinds of stays to help LLMs better capture the dependencies of people's mobility. 

\subsubsection{Capturing Long-term and Short-term Dependencies.}
In the context of the individuals' movements, long-term dependencies are associated with how people move in space over extended periods, often weeks to months or even years, which exhibit relatively stable patterns. For example, a person might always visit a certain workplace in the morning. In comparison, short-term dependencies appear more stochastic as they can be influenced by immediate and sometimes unpredictable factors like sudden changes in weather or an unplanned event, leading to variations in mobility patterns. To capture these dependencies, we propose to construct \textit{historical stays} as stays which are relatively more distant from the current time and span over a longer period, while \textit{context stays} are a few most recent stays that are less in number (as shown in Figure \ref{fig:dataorg}). The number of \textit{historical stays} and \textit{context stays} are \textit{M} and \textit{N}, respectively.

\subsubsection{Target Time Information for Time-Aware Prediction.}
Figure \ref{fig:dataorg} presents an illustration of the \textit{time-aware prediction} scenario, where the time information of the target stay is considered when predicting its corresponding place. This means that, instead of guessing the next place without knowing any temporal information (the \textit{time-unknown prediction} as shown in Figure \ref{fig:dataorg}), we predict which place the user will be at the next time $(st_{n+1},dow_{n+1})$. We argue that this is a more realistic implementation of the next location prediction problem because in real life, we would always like to know the situation at a specific time in the future. For example, a traffic forecasting system would predict the traffic condition at certain times in the future like 1 or 2 hours later; a user might ask an intelligent personal assistant (e.g., Alexa, Siri, etc.) a question like “Which leisure place do you recommend me to visit at around 3 pm this Saturday?”. 

\begin{figure}[t]
    \centering
    \includegraphics[width=1.0\columnwidth]{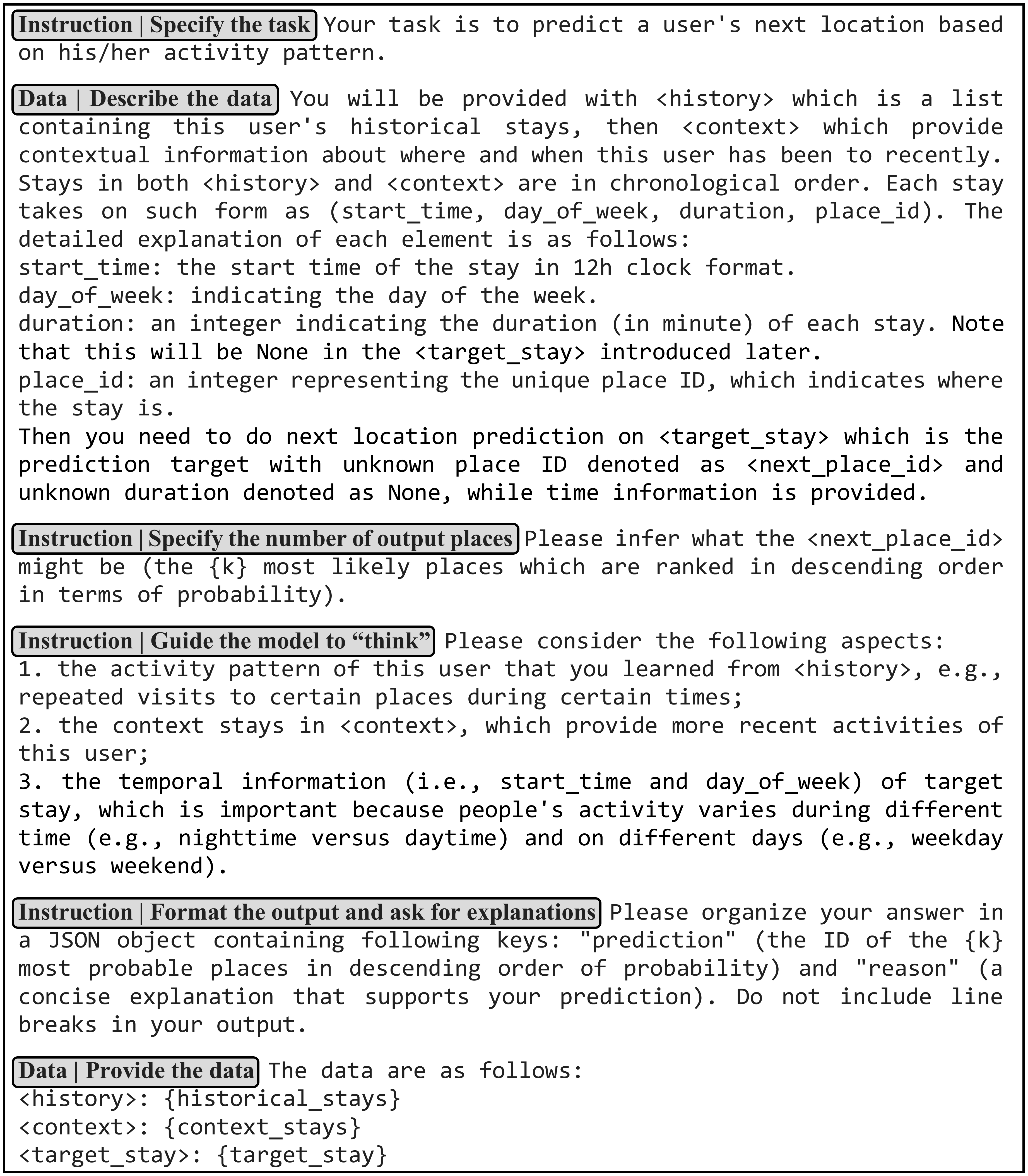}
    \caption{The prompt template used in \textit{LLM-Mob}. Each subpart is annotated by its function which is highlighted in a grey box. Please note that the annotations are not included in the prompt and are solely used for illustration.}
    \label{fig:prompt_wt}
\end{figure}

\subsection{Context-Inclusive Prompting}
Building upon existing prompting strategies like Chain-of-Thought (CoT) \cite{wei2022chain} and Plan-and-Solve (PS) \cite{wang2023plan}, we carefully develop 
context-inclusive prompts that incorporate relevant contextual information (such as the description for the data) to enhance next location prediction by LLMs. After iterative trials and refinements, we end up with the final prompt template shown in Figure \ref{fig:prompt_wt}. The template consists of two main aspects, i.e., instruction and data. Each aspect comprises several sub-parts which are annotated by their purpose. In the template, $<$history$>$, $<$context$>$ and $<$next\_place\_id$>$ correspond to \textit{historical stays}, \textit{context stays} and $pid_{n+1}$, respectively. Apart from general steps like \textit{Specify the task} and \textit{Provide the data}, we devise particular instructions and provide relevant context to make the most of the comprehension and reasoning capabilities of the LLMs, which are explained as follows.

\subsubsection{Describe the data.} We provide a detailed description of the three kinds of formatted data so that LLMs can easily digest the information in the context of human mobility prediction.

\subsubsection{Specify the Number of Output Places.}
We ask the model to output $k$ most probable places in descending order of probability so that we can evaluate the accuracy and cumulative gain up to rank position $k$ (refer to section \ref{sec:eval} for details on evaluation metrics). We hypothesise that the number of output places could potentially influence the model's performance, which is corroborated by the experiment results in Table \ref{tab:performance}.

\subsubsection{Guide the Model to “think”.}
To ensure that the model considers both long-term (historical activity pattern) and short-term (recent activity pattern) dependencies as well as the target time information in its reasoning process, we not only ask the model to consider these three aspects explicitly but also provide context to justify these considerations (e.g., inform the LLMs of the fact that people's activities vary during different times and on different days). We argue that this operation serves the same function as CoT \cite{wei2022chain}, i.e., guiding the model to “think” logically. However, CoT achieves this by providing demonstration examples, while we attain it using clear instructions and relevant context.

\subsubsection{Ask for Explanation.}
Moreover, we design our prompt to ask for both the prediction and the reason that supports the prediction. The intentions of this operation are: (1) Make the results interpretable, increasing the interpretability and reliability of the model and (2) Further enhance the reasoning ability of the model. We argue that asking for explanations is essentially the reasoning generation step proposed in PS prompting \cite{wang2023plan}.

\section{Experiments}
In this section, we evaluate the performance of our method both quantitatively and qualitatively. 

\subsection{Datasets and Preprocessing} 
We conduct extensive experiments on two types of human mobility datasets: A GNSS tracking dataset - Geolife \cite{zheng2010geolife} and a check-in dataset - Foursquare New York City (FSQ-NYC) \cite{yang2014modeling}. We strictly follow the data preprocessing steps proposed in \cite{hong2023context}, including filtering users with very few records, processing raw trajectories into stays and splitting the dataset into training and testing sets. 
The statistics information of the final datasets we used for evaluating our method are presented in Table \ref{data_stat}. 

\begin{table}[]
\begin{tabular}{lcc}
\toprule
                        & Geolife   & FSQ-NYC \\
\midrule
\# Users                & 45        & 535             \\
\# Days tracked  & 345 ± 413 &    84 ± 36             \\
\# Stays per user & 369 ± 456 &   178 ± 206              \\
\# Unique places per user   & 77 ± 108  &    34 ± 21             \\
\# Test samples         & 3459      & 12141 \\
\bottomrule
\end{tabular}
\caption{Basic statistics of the datasets (after preprocessing). The mean and standard deviation across users are reported.}
\label{data_stat}
\end{table}

\subsection{Experimental Settings}

\subsubsection{Implementation Details.}
The specific LLM employed in the experiment is GPT-3.5\footnote{https://platform.openai.com/docs/models/gpt-3-5}, which is one of the most advanced and widely used LLMs with open APIs\footnote{The specific version being used was gpt-3.5-turbo-0613.} and relatively lower cost. We set the temperature to 0 to avoid randomness in the output. Moreover, the length of historical stays \textit{M} and context stays \textit{N} are set as 40 and 5, respectively. These are empirical values that can be adapted to specific needs. 

\begin{table*}[]
\centering
\adjustbox{max width=\linewidth}{
\begin{tabular}{l|l|cccccc|ccc}
 \toprule[1pt]
\multirow{2}{*}{Dataset} & \multicolumn{1}{l|}{\multirow{2}{*}{Metric}} & \multicolumn{1}{c}{\multirow{2}{*}{1-MMC}} & \multicolumn{1}{c}{\multirow{2}{*}{LSTM}} & \multicolumn{1}{c}{\multirow{2}{*}{LSTM Attn}} & \multicolumn{1}{c}{\multirow{2}{*}{Deepmove}} & \multicolumn{1}{c}{\multirow{2}{*}{MobTcast}} & \multicolumn{1}{c|}{\multirow{2}{*}{MHSA}} & \multicolumn{3}{c}{\textbf{LLM-Mob}} \\
 & \multicolumn{1}{c|}{} & \multicolumn{1}{c}{} & \multicolumn{1}{c}{} & \multicolumn{1}{c}{} & \multicolumn{1}{c}{} & \multicolumn{1}{c}{} & \multicolumn{1}{c|}{} & $wot (k=10)$ & $wt (k=10)$ & $wt (k=1)$ \\
 \midrule
\multirow{5}{*}{Geolife} & Acc@1 (\%) & 24.1 & 28.4 & 29.8 & 26.1 & 29.5 & 31.4 & 33.0 & \underline{36.6} & \textbf{45.1} \\
 & Acc@5 (\%) & 38.1 & 55.8 & 54.6 & 54.2 & 51.3 & 56.4 & \textbf{82.9} & \underline{82.5} & - \\
 & Acc@10 (\%) & 39.5 & 59.1 & 58.2 & 58.7 & 56.2 & 60.8  & \underline{87.3} & \textbf{87.4} & -\\
 & Weighted F1 & 0.227 & 0.193 & 0.213 & 0.189 & 0.173 & 0.218 & 0.216 & \underline{0.259} & \textbf{0.404} \\
 & nDCG@10 & 0.327 & 0.447 & 0.450 & 0.426 & 0.434 & 0.465 & \underline{0.628} & \textbf{0.645} & -\\
\midrule
\multirow{5}{*}{\begin{tabular}[c]{@{}c@{}}FSQ-NYC    \end{tabular}} & Acc@1 (\%) & 16.0 & 16.9 & 17.0 & 19.3 & 20.2 & 20.2 & 23.2 & \underline{24.6} & \textbf{27.4}\\
 & Acc@5 (\%) & 32.5 & 39.0 & 39.4 & 43.3 & 45.9 & 47.0 & \underline{57.4} & \textbf{57.9} & -\\
 & Acc@10 (\%) & 36.4 & 47.1 & 47.5 & 52.3 & 55.5 & 57.3 & \underline{69.2} & \textbf{69.5} & -\\
 & Weighted F1 & 0.144 & 0.130 & 0.136 & 0.155 & 0.166 & 0.149 & 0.147 & \underline{0.172} & \textbf{0.221}\\
 & nDCG@10 & 0.266 & 0.312 & 0.314 & 0.349 & 0.368 & 0.376 & \underline{0.451} & \textbf{0.459} & -\\
 \bottomrule[1pt]
\end{tabular}
}
\caption{Comparison of predictive performance. $wt$ and $wot$ stand for $with$ and $without$ target time information,
respectively; $k=10$ and $k=1$ mean that the output length of the prediction results are set to 10 and 1 respectively. Numbers marked in \textbf{bold} and underline represent the best and second-best results, respectively.}
\label{tab:performance}
\end{table*}

\subsubsection{Baselines.} We compare our model with classical prediction models and recently published state-of-the-art deep learning models in the next location prediction domain. 

\begin{itemize}
\item 1-MMC. First-order Mobility Markov Chain \cite{gambs2012next}, a classical location prediction model which assumes the Markovian property for individual location visits. 
\item  LSTM. A classic sequence modelling neural architecture that has shown great performance for the next location prediction \cite{krishna2018lstm,solomon2021analyzing}.
\item LSTM-Attn. An LSTM with a (masked) self-attention between the hidden states \cite{li2020hierarchical}. The attention results are combined with the actual output of the LSTM through a small feedforward network.
\item DeepMove. A framework comprising two separate recurrent networks; one for learning the periodicity from historical visits and the other for mining transition patterns from the current trajectory \cite{feng2018deepmove}.
\item MobTcast. Based on the transformer encoder network, MobTCast leverages temporal, semantic, social, and geographical contexts in the history place sequence to predict the next place \cite{xue2021mobtcast}.
\item MHSA. Multi-Head Self-Attentional (MHSA) neural network that builds on transformer architecture, considering historical location visits, visit time, activity duration, as well as their surrounding land use functions, to infer an individual’s next location \cite{hong2023context}.
\end{itemize}

\subsubsection{Evaluation Metrics.}
\label{sec:eval}
We use the following commonly used metrics to quantify the predictive performance of compared models: 

\begin{itemize}
    \item \textbf{\textit{Accuracy}}. 
    Predictions are sorted in descending order based on their probability of being the next location, and Acc@k measures the proportion of times the ground truth location appears within the top-k predictions. We report Acc@1, Acc@5, and Acc@10 for comparison.
    \item \textbf{\textit{Weighted F1}}. Individual's visits to locations are highly unbalanced, with some locations occurring more often than others. We use the F1 score weighted by the number of visits to emphasise the model’s performance in the more important locations.
    \item \textbf{\textit{nDCG@k}}. Normalized discounted cumulative gain (with rank position $k$) measures the quality of the prediction vector by the ratio between the discounted cumulative gain (DCG) and the ideal discounted cumulative gain (IDCG). The calculation of \textit{nDCG@k} is given below:
    \begin{equation}
        nDCG@k= \frac{DCG_k}{IDCG_k}, 
    \end{equation}
    \begin{equation}
        DCG_k=\sum_{j=1}^{k} \frac{r_j}{log_2(j+1)},
    \end{equation}
    where $r_j$ denotes the relevance value at position $j$. In the context of location prediction, $r_j \in \{0, 1\}$, and $r_j=1$ if and only if the $j$-th item in the ranked prediction vector corresponds to the ground truth next location. In our experiment, we report the average \textit{nDCG@10} over all test samples.
\end{itemize}

\subsection{Accuracy Comparison}
The predictive performance of all considered methods is presented in Table \ref{tab:performance}. The results show that our model significantly outperforms all baseline models in terms of all the metrics. 
Comparing model variants \textit{wot} and \textit{wt} under the same output length $k=10$, we can see that incorporating time information of the target stay enhances the performance.
When asking the model to output the most likely place instead of top 10 probable places, the \textit{Acc@1} and \textit{F1 score} improve further by a large margin. The results demonstrate that \textit{LLM-Mob} is a promising approach for predicting human mobility.

\subsection{Ablation Study}
To evaluate the effects of the three components proposed in the data formatting step, we build three variants of the full model (\textit{LLM-Mob}), i.e., \textit{NoHistory}, \textit{NoContext} and \textit{NoTime}, which exclude historical stays, context stays and target stay time information as well as relevant contents in the prompt, respectively. To evaluate the effects of specific prompting strategies, two more variants are designed. \textit{NoGuide} removes the “\textit{guide the model to ‘think’}” part in the prompt; \textit{NoReason} deletes the requirement for \textit{LLM-Mob} to output reasons that support the prediction. 

\begin{figure}[t]
    \centering
    \includegraphics[width=0.8\columnwidth]{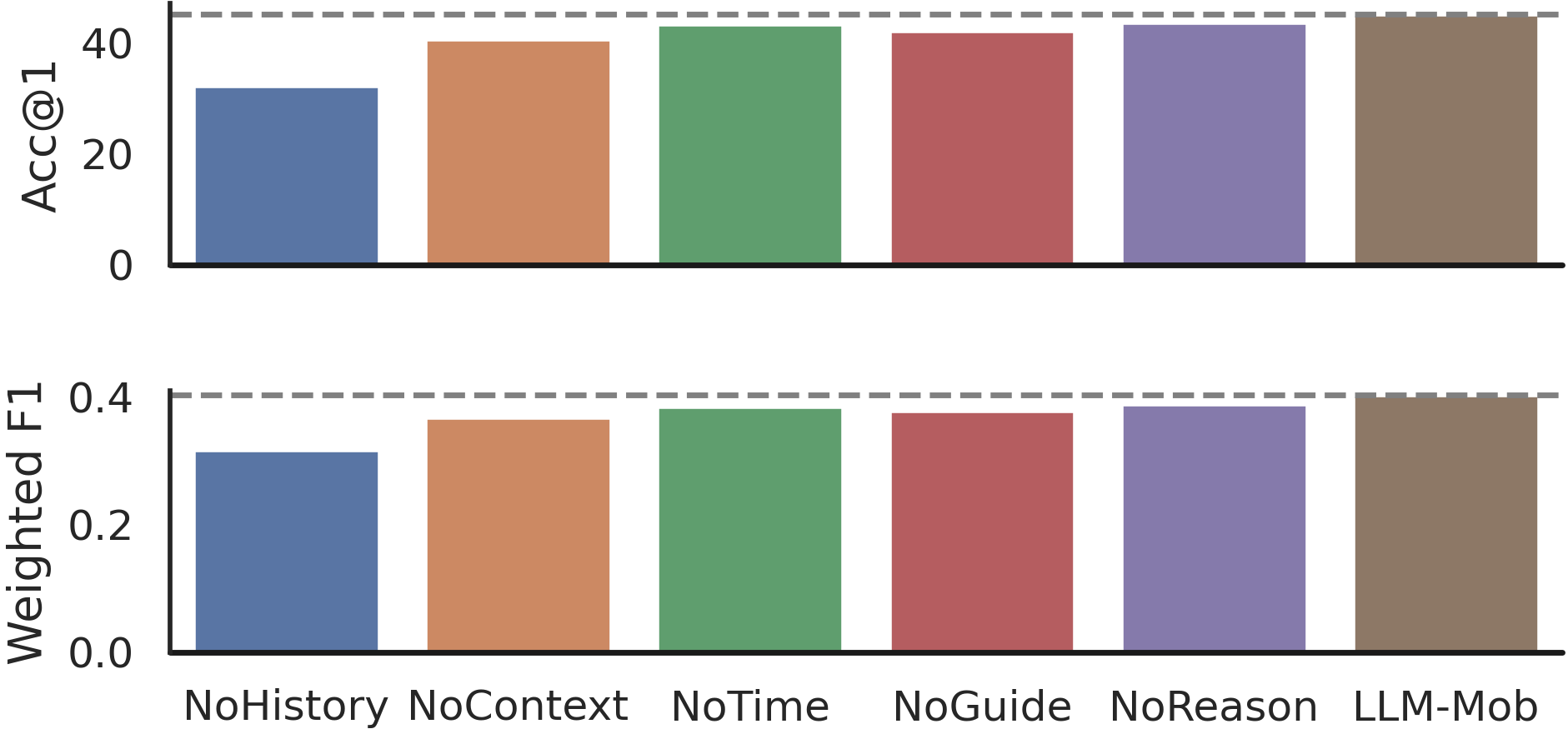}
    \caption{Ablation study on Geolife}
    \label{fig:ablation}
\end{figure}

We perform an ablation study with these five variants under $k=1$ time-aware prediction setting and report the performance on Geolife dataset in Figure \ref{fig:ablation}. We can observe that the full model consistently outperforms all the variants, indicating that each designed component has a positive effect on performance improvement. Moreover, the performance drop with \textit{NoHistory} is particularly higher than others, further underscoring the importance of considering long-term dependencies in predicting people's location.

\subsection{Case Study}
\label{sec:interp}
We select and analyse some example results from Geolife (as shown in Table \ref{tab:casestudy}) to gain a deeper understanding of \textit{LLM-Mob}’s reasoning ability and interpretability.

\begin{table}[]
\begin{tabular}{lll}
\toprule
Experiment   settings                           & Acc@1 (\%)       & Weighted F1  \\ \midrule
Temperature=0.5             & 44.6 (0.98) & 0.398 (0.020) \\ 
Paraphrased prompts          & 44.9 (0.56) & 0.406 (0.012) \\ 
GPT-4 (temperature=0)                 & 45.3        & 0.405       \\ \bottomrule
\end{tabular}
\caption{Additional experiments. The numbers in brackets are standard deviations over five runs.}
\label{tab:robust}
\end{table}

\begin{table*}[!ht]
\centering
\small\setlength{\tabcolsep}{4pt} 
\resizebox{0.8\linewidth}{!}{
\renewcommand{\baselinestretch}{0.8}\selectfont
\begin{tabular}{p{0.1\textwidth} p{0.1\textwidth} p{0.8\textwidth}}
\toprule[1pt]
\multicolumn{3}{l}{\textbf{Case 1: the 5th test sample of user 1.}} \\
\midrule
\multicolumn{2}{l}{\textbf{Ground truth: }} & \textcolor{green}{1} \\
\multicolumn{2}{l}{\textbf{Target time information: }} & (15:13, Friday) \\
\midrule
\textbf{MHSA} & Prediction: & [17, 10, 4, 22, 20, 18, 9, 123, 59, 84] \\
& Reason: &  None
\\
\textbf{LLM-Mob ($k=10$)} & Prediction: & [17, 10, \textcolor{green}{1}, 18, 4, 39, 42, 38, 7, 21] \\
& Reason: & Based on the user's historical stays, they have frequently visited places 17, 10, and 1. Additionally, the user has recently been to place 10 in the context stays. Considering the temporal information of the target stay, which is on a Friday afternoon, it is likely that the user will visit a place for leisure or social activities. Therefore, places 17, 10, and 1 are the most probable options. Other options include places 18, 4, 39, 42, 38, 7, and 21, which the user has visited before and are also popular leisure or social destinations.
\\
\textbf{LLM-Mob ($k=1$)} & Prediction: & \textcolor{green}{1} \\
& Reason: & Based on the user's history, they have visited place 17 multiple times during weekdays and have stayed for a considerable amount of time. Additionally, the user's most recent context stays have been at places 1 and 10. Considering the time of the target stay (Friday afternoon), it is likely that the user will visit a place for a shorter duration, possibly for a quick errand or lunch. \textcolor{blue}{Therefore, place 1, which is a restaurant, is a likely next destination for the user.}
\\
\midrule
\midrule
\multicolumn{3}{l}{\textbf{Case 2: the 5th test sample of user 13.}} \\
\midrule
\multicolumn{2}{l}{\textbf{Ground truth: }} & \textcolor{green}{9} \\
\multicolumn{2}{l}{\textbf{Target time information: }} & (23:50, Tuesday) \\
\midrule
\textbf{MHSA} & Prediction: & [\textcolor{green}{9}, 883, 225, 332, 445, 10, 1147, 344, 614, 685] \\
& Reason: &  None
\\
\textbf{LLM-Mob ($k=10$)} & Prediction: & [445, \textcolor{green}{9}, 444, 335, 448, 447, 446, 1, 444, 443] \\
& Reason: & Based on the user's historical stays, they have a pattern of visiting places 445, 9, and 444 frequently. Additionally, the contextual information shows that the user recently visited place 445 and 444. Considering the temporal information of the target stay, which is nighttime on a weekday, it is likely that the user will visit one of these familiar places again. Other possible options include places 335, 448, 447, 446, 1, and 443, which have been visited by the user before and are open during nighttime hours.
\\
\textbf{LLM-Mob ($k=1$)} & Prediction: & \textcolor{green}{9} \\
& Reason: & Based on the user's history, they have visited place ID 9 multiple times during nighttime hours (e.g., 11:59 PM on Sunday and 11:53 PM on Tuesday). Additionally, the target stay is also during nighttime hours on a weekday, which further supports the likelihood of the user visiting place ID 9.
\\
\bottomrule[1pt]
\end{tabular}
}
\caption{Case study of test examples. Ground truth and the correct predictions are marked in green. The “hallucination” problem is marked in blue.}
\label{tab:casestudy}
\end{table*}

From case 1, we can see that \textit{LLM-Mob} correctly predicted place 1 as the next location under $k=1$ setting, and it also predicted it to be among the “top 10” list of predictions under $k=10$ setting, while MHSA failed to do so. \textit{LLM-Mob} provided detailed and rational reasons for the predictions it made, while MHSA was unable to explain the results. Moreover, our model can take the target time information (if available) into account to make more accurate prediction.

However, \textit{LLM-Mob} also has its weaknesses. From case 2, we can see that \textit{LLM-Mob} generated duplicated place IDs within the top 10 list, and we argue this was mainly due to the limited number of unique places user 13 had been to. Also, it is not as flexible as MHSA, as can be seen in the $k=10$ setting where it assumed the most frequently visited place (445 in this case) as the most likely next location, whereas MHSA could give more varied responses based on the input. There are also “hallucination” problems seen in the reasons generated by \textit{LLM-Mob}, which is discussed in section \ref{sec:limit}.

\begin{table*}[!ht]
\centering
\small\setlength{\tabcolsep}{4pt} 
\resizebox{0.8\linewidth}{!}{
\renewcommand{\baselinestretch}{0.8}\selectfont
\begin{tabular}{p{0.1\textwidth} p{0.9\textwidth}}
\toprule[1pt]
\multicolumn{2}{l}{\textbf{Task 1: Converting a structured numerical input to its natural language description.}} \\
\midrule
\textbf{Prompt: } & Your task is to convert a raw $<$input$>$ containing human mobility information to its natural language description. The $<$input$>$ is a list containing a user's historical stays...Omitted description for stay...
Please be concise and only describe when and where the user went and how long he/she stayed there.
$<$input$>$: [(4, 289, 127, 335), (5, 614, 1332, 444), (6, 599, 1398, 444), ...]
\\
\textbf{Answer: } & The user's historical stays are as follows:
On Thursday, the user stayed at place ID 335 from 4:49 AM to 7:36 AM for a duration of 2 hours and 7 minutes.
On Friday, the user stayed at place ID 444 from 10:14 AM to 8:26 PM for a duration of 22 hours and 18 minutes.
On Saturday, the user stayed at place ID 444 from 9:59 AM to 9:37 PM for a duration of 23 hours and 38 minutes. …
\\
\midrule
\midrule
\multicolumn{2}{l}{\textbf{Task 2: Summarising the activity patterns.}} \\
\midrule
\textbf{Prompt: } & Based on the information provided, please summarize the activity patterns of this user. Please consider both place and temporal information like weekday and start time.
\\
\textbf{Answer: } & Based on the information provided, we can summarize the activity patterns of this user as follows:
Weekday Preference: The user's most frequent stays occurred on Mondays and Tuesdays, with a total of 7 stays on these days. They also had significant activity on Thursdays and Fridays, with 6 stays on each of these days.
Morning Activity: The user tended to start their stays in the morning hours, with most stays starting between 6:00 AM and 11:00 AM.
Diverse Places: The user visited multiple places, including IDs 335, 444, 445, 446, 447, 448, and 9. They had the highest number of stays at place ID 445, with a total of 12 stays.
Long Stays: The user had some longer stays, with durations ranging from a few minutes to as long as 23 hours and 18 minutes. The average duration of their stays was relatively high, indicating they spent considerable time at each place.
Weekend Activity: The user had a notable stay on Saturday, but their activity was relatively lower on Sundays, with only one stay recorded.
Overall, the user exhibited a regular pattern of activity, with preferences for specific weekdays and morning hours. They frequented multiple places, spending significant time at each location.
\\
\midrule
\midrule
\multicolumn{2}{l}{\textbf{Task 3: Inferring the home and workplace.}} \\
\midrule
\textbf{Prompt: } & Based on your analysis, can you further infer the home and workplace of this user?
\\
\textbf{Answer: } & Workplace Inference:
We can observe that the user frequently stayed at place ID 445 on weekdays, with a total of 7 stays during Monday to Friday. This place ID occurs consistently during morning hours, suggesting a potential workplace.
Home Inference:
There are no specific stays that occur consistently during evenings and nights, indicating that determining the home location solely based on the provided data is challenging. Therefore, based on the information provided, we can speculate that place ID 445 could potentially be the user's workplace. However, without additional data or specific knowledge about the user's routines and preferences, it is difficult to determine the exact home location.
It's important to note that this inference is based on assumptions and may not accurately represent the user's actual home and workplace.
\\
\bottomrule[1pt]
\end{tabular}
}
\caption{Test results from 3 progressive tasks.}
\label{tab:discuss}
\end{table*}

\subsection{Robustness Test}

We conduct additional experiments (Table \ref{tab:robust}) to investigate the robustness of \textit{LLM-Mob} concerning (1) non-zero temperature; (2) changing of prompts;  (3) using a different LLM. All these experiments are performed under $k=1$ time-aware prediction setting on Geolife dataset. Overall, the results show no significant variations in predictive performance, indicating the robustness of our method. We anticipate that higher temperature and greater changes of prompts would bring greater uncertainties. Therefore, in real-world applications that favour stable performance, lower temperature and consistent prompts would be preferred.

\section{Discussion}
\subsection{Why does \textit{LLM-Mob} perform so well?}
In this section, 
we test LLM’s performance on three progressive tasks to demonstrate different aspects of its capabilities that we argue are essential to successful human mobility prediction. 
The results presented in Table \ref{tab:discuss} show that the LLM has done a great job in all three tasks, i.e., generating the natural language description of a numerical and structured input, summarising the activity patterns of the user, and making inference about the home and workplace of the user. This has shown three major capabilities:
(1) The ability to understand not only natural language but also structured numerical input like a list of numbers representing trajectories.
(2) The capability of summarising historical mobility/activity patterns so that the model can effectively draw from past information to inform the future situation.
(3) The powerful reasoning ability, meaning that the model can “think” like a human and make rational inferences.
Apart from the inherent capabilities of LLM, the success of \textit{LLM-Mob} also lies in the effective formatting of data and practical prompt engineering where the prompt has been carefully designed and improved through iterative testing. 

\subsection{Limitations}
\label{sec:limit}
There are several limitations: (1) Efficiency issue. Currently, our method calls OpenAI API for each test sample independently, which is inefficient and is hard to apply to large-scale predictions. (2) Hallucination \cite{bang2023multitask}. This is a common problem faced by LLMs. As can be seen from the statement marked in blue in Table \ref{tab:casestudy}, the model stated that place 1 is a restaurant, which is fabricated and can be misleading to the user. (3) Limitations from the proprietary LLMs. 
Firstly, calling the OpenAI API costs money, and it can be a lot when the data volume is big. Moreover, OpenAI is constantly updating the GPT model family, resulting in performance drift of the newest models \cite{chen2023chatgpt}. Consequently, the prompts that perform well on old models may not work on new ones, requiring extra work on prompt engineering. 
We argue that all these limitations necessitate the training of open-source LLMs (or rather, foundation models) fine-tuned for modelling human mobility, which can not only have better predictive performance, but also give researchers and practitioners full control of their model that can be adapted to their needs.

\section{Conclusions}
In this study, we established a novel framework that adopted LLMs as its backbone for human mobility prediction. In this framework, mobility data were formatted to account for both long-term and short-term dependencies and to facilitate time-aware predictions. Furthermore, context-inclusive prompts were formed and engineered to enable the reasoning process performed by LLMs to generate accurate predictions and logical explanations. Extensive  experiments were conducted on two real-world publicly-available human mobility datasets and the results demonstrated state-of-the-art predictive performance of the proposed model as well as its superior interpretability.

However, it should be noted that the predictive performance of our method still has room for improvement and limitations like hallucination should be further addressed. We hope the findings in this work could shed light on human mobility research domain and provide insights into future applications of LLMs.

\appendix
\section*{Ethical Statement}

This work used anonymised open data that has no personally identifiable information. Whether the real-world applications of this work cause privacy issues depends on the specific providers of LLMs. We expect the potential risk to be minimal, as all proprietary LLMs abide by strict privacy policies \footnote{https://openai.com/policies/privacy-policy}. 
Moreover, in use cases like travel demand forecasting, personal data will be safely discarded after the aggregation of mobility patterns to specified regions, maximising the preservation of individual's privacy.

\bibliographystyle{named}
\bibliography{ijcai24}

\end{document}